\documentclass[conference]{IEEEtran}

\usepackage{multicol}
\usepackage{multirow}
\usepackage{caption}
\usepackage{subcaption}
\usepackage{textcomp}
\usepackage{array}

\usepackage[pdftex]{graphicx}
\usepackage{url}

\begin{document}

\title{Database of iris images acquired in the presence of ocular pathologies and assessment of iris recognition reliability for disease-affected eyes}

\author{\IEEEauthorblockN{Mateusz Trokielewicz$^{\dag,\ddag}$, Adam Czajka$^{\ddag,\dag}$}
\IEEEauthorblockA{$^{\dag}$Biometrics Laboratory\\Research and Academic Computer Network (NASK)\\
Wawozowa 18, 02-796 Warsaw, Poland\\
$^{\ddag}$Institute of Control and Computation Engineering\\Warsaw University of Technology\\
Nowowiejska 15/19, 00-665 Warsaw, Poland\\
contact: \url{{mateusz.trokielewicz,adam.czajka}@nask.pl}}
\and
\IEEEauthorblockN{Piotr Maciejewicz$^{\star}$}
\IEEEauthorblockA{$^{\star}$Department of Ophthalmology,\\
Medical University of Warsaw\\
Lindleya 4, 02-005 Warsaw, Poland\\
contact: \url{piotr.maciejewicz@wum.edu.pl}}}

\IEEEoverridecommandlockouts
\IEEEpubid{\makebox[\columnwidth]{Manuscript accepted for publication at the IEEE CYBCONF2015, Gdynia, Poland} 
\hspace{\columnsep}\makebox[\columnwidth]{ }}

\maketitle

\begin{abstract}
This paper presents a database of iris images collected from disease affected eyes and an analysis related to the influence of ocular diseases on iris recognition reliability. For that purpose we have collected a database of iris images acquired for 91 different eyes during routine ophthalmology visits. This collection gathers samples for healthy eyes as well as those with various eye pathologies, including cataract, acute glaucoma, posterior and anterior synechiae, retinal detachment, rubeosis iridis, corneal vascularization, corneal grafting, iris damage and atrophy and corneal ulcers, haze or opacities. To our best knowledge this is the first database of such kind that will be made publicly available. In the analysis the data were divided into five groups of samples presenting similar anticipated impact on iris recognition: 1) healthy (no impact), 2) unaffected, clear iris (although the illness was detected), 3) geometrically distorted irides, 4) distorted iris tissue and 5) obstructed iris tissue. Three different iris recognition methods (MIRLIN, VeriEye and OSIRIS) were then used to find differences in average genuine and impostor comparison scores calculated for healthy eyes and those impacted by a disease. Specifically, we obtained significantly worse genuine comparison scores for all iris matchers and all disease-affected eyes when compared to a group of healthy eyes, what have a high potential of impacting false non-match rate.
\end{abstract}
\begin{keywords}iris recognition, eye conditions, iris image databases, performance evaluation.\end{keywords}

\section{Introduction}

Iris recognition is envisioned as one of the most accurate biometric authentication method and evaluated in the most impressive biometric projects worldwide, including generation of the unique identification numbers for the population of India, ADHAAR \cite{AADHAAR}, or Canadian border control system CANPASS \cite{CBSA}. However, the eye and iris -- like any other human organs -- may suffer from various diseases that may influence the biometric processes. There are various interesting questions in case large-scale systems, in particular: Which eye conditions impact the reliability of the iris recognition? To what extent the accuracy is deteriorated? Are there any countermeasures that we may apply? This paper contributes to answering at least two first questions, proposing a database\footnote{BioBase-Disease-Iris v1.0 is publicly available for research and non-commercial use. See \url{http://zbum.ia.pw.edu.pl/EN/node/46} for further details.} of images (detailed in Sec. \ref{sec:Database}) of eyes affected by numerous diseases (described in Sec. \ref{section:medical}) as well as an experimental study based on three different iris recognition methods (briefly characterized in Sec. \ref{section:methods}) applied for eyes demonstrating a few types of visual impairments (Sec. \ref{sec:expmethodology}). In particular we show significant decrease in similarity among disease-affected samples of the same eyes (Sec. \ref{sec:Results}), even if no visible changes can be distinguished under NIR imaging.

\section{Medical disorders affecting the iris}
\label{section:medical}

The following section briefly describes several types of ocular diseases that were encountered throughout the process of creating our database. Both symptoms and possible influences that they may have on the reliability of iris recognition are discussed.

\begin{figure*}[!t]
\centering
\includegraphics[width=\textwidth]{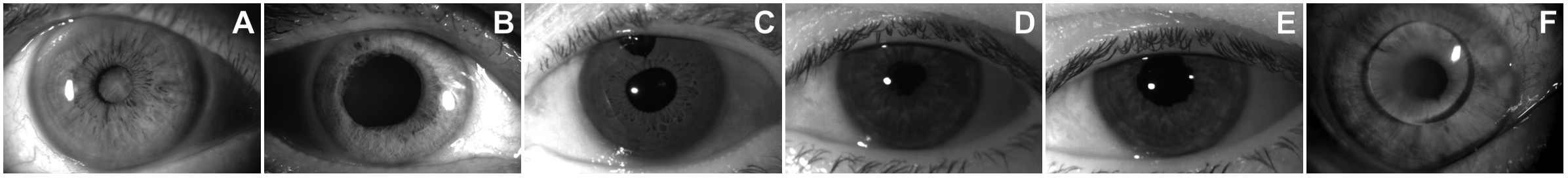}
\caption{Samples representing ocular pathologies or conditions discussed in Section \ref{section:medical}: cataract (A), acute glaucoma (B), iridectomy (C), synechiae before and after lens replacement (D, E), silica oil in the anterior chamber of the eye (F).}
\label{fig:diseases}
\end{figure*}

One of the most common ophthalmic disorders worldwide is \textbf{cataract}. It causes partial or total blurring of the eye lens, resulting in impaired and dimmed vision, Fig. \ref{fig:diseases}A. This happens due to the fact that light is unable to properly penetrate the opacified lens and focus onto the retina. Here, negative consequences for biometric recognition may arise from the grayish pupillary area contributing to the iris segmentation errors when using certain algorithms deploying gradient information between the pupil and the iris. Moreover, cataract is often accompanied by various conditions, such as acute glaucoma, anterior and posterior synechiae, iris atrophy, pseudoexfoliation syndrome and more, all of them capable of distorting the iris or other eye structures. Cataract treatment usually incorporates lens replacement with the implantation of artificial lens, with a non-zero chance of inflicting damage to the iris tissue.

The \textbf{acute glaucoma} is a condition bound to happen when the space between the iris and the cornea closes completely on the outer iris boundary, blocking the flow of the aqueous humor through the trabecular meshwork and causing a sudden increase in the intraocular pressure, that -- if not treated -- may lead to an immediate and complete loss of vision due to atrophy of retina ganglion cells \cite{GlaucomaWHO}. This increased pressure, imposing force against the iris from inside of the eyeball, is capable of causing a certain \emph{flattening} of the iris tissue and a distortion in the pupil shape, Fig. \ref{fig:diseases}B. Treatment incorporates either using a laser to make a small, point-shaped incision in the iris, or filtration surgery with removing part of the trabecular meshwork (typically triangle-shaped and located in upper part of the iris) to perform drainage of aqueous humor from the eye, Fig. \ref{fig:diseases}C.

 \textbf{Posterior and anterior synechiae} occur when the iris becomes partially attached to the lens or to the cornea. This can alter the appearance of the pupil, causing it to deviate from its usual circular shape, Fig. \ref{fig:diseases}D. Also, when the iris adheres to the lens it may render its surface brighter than normally, causing trouble similar to those associated with the cataract. Synechiae accompanying cataract, when the latter is treated with the lens replacement, can also leave a permanent deformation of the pupil, even though there is no longer any adherence between the iris and newly implanted lens, Fig. \ref{fig:diseases}E.
  
 \textbf{Retinal detachment} occurs when a retina detaches from the layers below it causing a visual impairment that may lead to a complete blindness if left unattended. The condition itself has little impact on the look of the iris, however, treatment usually applied in such cases incorporates filling the eyeball with silica oil to attach the retina with the back of the eyeball cavity. Then a laser is used to finally combine the retina with the wall of the eyeball. The applied oil can in certain cases make its way back from the inside of the eyeball, creating an obstruction, Fig. \ref{fig:diseases}F.

Numerous other eye diseases and conditions, happening more or less frequently, have a capability to affect iris recognition. Those existing in the discussed dataset include \textbf{rubeosis iridis} (pathological vascularization on the iris surface due to the growth factors released by an ischemic retina \cite{Rubeosis}), \textbf{corneal vascularization} with minute blood vessels present in the cornea and preventing the light from properly penetrating it, \textbf{corneal ulcers, haze or opacities} of different origin but with consequences similar to those of angiogenesis, \textbf{corneal grafting} with visible sutures, various types of \textbf{iris damage and atrophy} and more. Appropriate characterization of each sample collected in the database is given in the associated metadata. 

\section{Related work}

Roizenblatt \emph{et al.} \cite{Roizenblatt} were the first to analyze how eye pathologies influence the performance of biometric iris recognition, investigating 55 cataract patients and reporting an FNMR of 11\% when comparing post-surgery iris images to those obtained beforehand. At the same time researchers point to a correlation between similarity score degradation and an increase in a score denoting visible changes to the iris, such as pupil ovalization, iris atrophy and depigmentation. Cataract surgery and mydriatics influence is also studied by Dhir \emph{et al.} \cite{Dhir}, who report no false non-matches of images of irides after the treatment when compared to those taken before the surgery. However, it is worth noticing that a truncated dataset has been used, \emph{i.e.}, each eye with visible damage caused by the medical procedure had been excluded from it. Authors also investigate a possible decrease in similarity scores caused by excessive pupil dilation due to the use of mydriatics (pupil-dilating drugs common in ophthalmological practice), noting a FNMR of 13\% when comparing images of eyes with pupils pharmaceutically dilated to those obtained before mydriatics instillation. Seyeddain \emph{et al.} \cite{Seyeddain2014} observed a 5\% FNMR when comparing after-surgery iris images to pre-surgery ones. Also, an FNMR of 12\% is found when comparing images representing drug-dilated pupils to those of the same eyes, but without this kind of medication. Different aspect of issues caused by cataract is reported by Trokielewicz \emph{et al.} \cite{TrokielewiczWilga2014}, who provide an account on whether cataract-affected eyes perform generally worse when employed for iris recognition purposes, in situations when no surgical treatment has yet been applied. Authors find a statistically significant decrease in average genuine comparison score when comparing scores obtained from cataract-affected eyes to those calculated for healthy eyes, reaching 175\% of decrease for a selected commercial iris recognition methodology. Moreover, an analysis of possible faulty iris segmentation being a cause of this behavior leads to a conclusion that performance degradation is not explicitly connected with erroneous segmentation results. Yuan \emph{et al.} \cite{Yuan} provide an account on possible connection between laser-assisted refraction correction surgeries and a decreased biometric reliability. However, out of 14 eyes with surgery performed, only one had not been recognized afterwards and a severe deviation from pupil's circularity was present, most likely being responsible for this failure. Borgen \emph{et al.} \cite{Borgen} perform digital alterations on the selected images from the UBIRIS dataset so that they resemble several conditions, such as corneal diseases, angiogenesis, tumors, iridectomy and iris depigmentation. False non-match rates reaching as high as 87\% are reported when comparing the altered images to the same files but before the tampering.

The only approach to examining biometric performance in relation to various eye pathologies appears in a work by Aslam \emph{et al.} \cite{Aslam}, who attempt to assess the extent of method reliability degradation after carrying out a treatment to eyes suffering from cornea and sclera pathologies, glaucoma, conjunctivitis and other. Tested recognition methodology was found to be resilient for most diseases except in some cases of iritis, in which an FNMR of 21\% was obtained. McConnon \emph{et al.} \cite{McConnon2012} also examine a dataset consisting of samples epitomizing multiple pathologies, but the image segmentation stage is only performed and no sample matching technology is employed. Authors report that the segmentation results vary by more than 2 pixels from the manually obtained ground truth in about half of the images, however, one should be aware of the fact that the images involved had been obtained using an ophthalmoscope equipped with a color camera and thus are not well suited for the purpose of biometric identification, which can cause the results to be somehow biased.

\section{Database}
\label{sec:Database}
\subsection{Principles of building the database}

Database described in this paper consists of images collected during routine ophthalmology visits of patients of the Department of Ophthalmology of the Medical University of Warsaw. All patients have been provided with detailed information about this study and a written consent has been signed by all volunteers.

Our dataset consists of iris images acquired under near-infrared (NIR) illumination and compliant with the ISO/IEC iris image quality standards \cite{ISO}.  Furthermore, we have also collected samples in visible light in those cases, where visual inspection conducted by an ophthalmologist would reveal significant alterations to eye structures, most notably to the iris itself. This is to assert whether certain conditions that disturb the appearance of the iris in visible light illumination also reveal themselves when NIR illumination is used, as some studies show that certain types of pathologies, such as corneal clouding, may obstruct the iris in visible light, but not in NIR \cite{Aslam}.

The data collection process lasted for approx. 8 months. During the enrollment, a typical ophthalmology examination was performed by an ophthalmology specialist. Each patient was given a unique ID number and a \textbf{visit} was created in the system. Apart from gathering typical information (the case and treatment details: \textbf{case description}), image acquisition was performed. At least six NIR and six visible light photographs of each eye have been captured (\textbf{images}). Notably, each sample is considered a \emph{separate attempt, i.e.}, the patient had to lift his/her head from the chin and forehead rests after each capture attempt. This is done to deliberately introduce a certain level of noise in the intra-session sample sets. This strategy was then repeated for every future visit, each of which has a separate visit metadata, \emph{e.g.}, to distinguish between pre- and post-treatment image subsets. Each eye is treated independently in this collection. Summarizing, the data for each eye comprises \textbf{visits} that include \textbf{case descriptions} and \textbf{images}.

\subsection{Equipment used}
To obtain good quality NIR images we employed a commercial iris recognition camera, the \textbf{IrisGuard AD100}, that generates images in a standard VGA resolution (640$\times$480 pixels) while also meeting the quality requirements defined in the aforementioned ISO/IEC documents. To speed up the data collection the device was placed on a stand with fixed chin and forehead rests. This makes the setup instantly familiar for both the ophthalmologist and the patient while at the same time reduces eye movement and thus the possible motion blur in the outcome images.

As for the visible light (\emph{i.e.}, color) images, two different cameras were used. \textbf{Canon EOS 1000D} dSLR camera with bundled lens (EF-S 18-55 mm f/3.5-5.6 IS) was equipped with a Raynox DCR-250 macro lens converter to enhance close-up image quality and a macrophotography LED ring flashlight to provide sufficient lighting. The Canon camera produces 10 megapixel images with JPEG compression. This setup was then established on a stand similar to the one used with the AD100 device. In some cases the eye was also imaged with a digital camera attached to a professional ophthalmoscope, the \textbf{Topcon DC3 slit-lamp camera}, to provide more detailed view of a particular region of interest. This device is capable of producing 8 megapixel, JPEG-compressed images.

\subsection{Data censoring}
Following the process of collecting samples, each of them had to be carefully evaluated to exclude all images not compliant to the requirements defined in the ISO/IEC standards. This is to make sure that experimental results are not obscured by poor quality of images, with flaws such as not containing the iris at all or showing less than 70\% of the iris, severely blurred images, images showing pupil-to-iris diameter ratio falling outside of the [0.2, 0.7] range or images with 'gaze-away' eyes.

\subsection{Database summary}
After censoring the resulting dataset consisted of 603 NIR-illuminated images of 91 different eyes (later on referred to as the NIR\_AD100 subset) and 222 color images of 25 eyes (the VIS\_CANON and the VIS\_TOPCON subsets, combined). Table \ref{database_summary} summarizes the database in respect to these three subsets. 

\begin{table}[!ht]
\renewcommand{\arraystretch}{1.1}

\caption{Database subsets summary.}
\label{database_summary}
\centering
\begin{tabular}{|c|c|c|c|c|}
\hline
\textbf{Subset} & \textbf{Device} & \textbf{Classes (eyes)} & \textbf{Images} & \textbf{Per class} \\
\hline
\hline
NIR\_AD100 & IrisGuard AD100 & 91 & 603 & (2, 26)\\
\hline
VIS\_CANON & Canon EOS 1000D & 22 & 138 & (0, 10)\\
\hline
VIS\_TOPCON & Topcon DC3 & 25 & 84 & (0, 7)\\
\hline
\end{tabular}
\end{table}

\section{Iris recognition methods used in this work}
\label{section:methods}

In this study three well-established iris recognition methods were used: two commercially available (MIRLIN and VeriEye) and one open-source solution (OSIRIS). All three methods were used as 'black boxes', \emph{i.e.}, we did not analyze sub-processes such as iris segmentation or encoding, and only the comparison results were taken into account. 

The first method, {\bf MIRLIN} (Monro Iris Recognition Library) \cite{MIRLIN}, derives the iris features from the zero-crossings of the differences between Discrete Cosine Transform (DCT) calculated in rectangular iris image subregions \cite{MIRLINpaper}. The coding method yields binary iris codes, thus the comparison incorporates a normalized Hamming Distance. The lower the score, the better the match (the smaller distance between samples).

The second method, {\bf VeriEye} offered by Neurotechnology \cite{VeriEye}, employs a proprietary and not published iris coding methodology. The manufacturer claims a correct off-axis iris segmentation with the use of active shape modeling, in contrast to typical circular approximation of the iris boundaries. VeriEye was tested for a few standard iris image databases, it was used in the NIST IREX project and presents good accuracy. The resulting score corresponds to the similarity of samples, \emph{i.e.}, the higher the score, the better the match.

The third method, {\bf OSIRIS}, is an open source software following the Daugman's idea of using Gabor filters to enhance individual iris features and diminish all the remaining properties \cite{OSIRIS}. After image filtering each of the resulting composite vectors is quantized to two bits. Since the iris image is normalized before filtering, the binary code has identical length and the structure for each iris. Hence, as in MIRLIN matcher, a normalized Hamming Distance is used to calculate the distance between feature vectors, \emph{i.e.}, lower comparison scores denote more similar samples.

\section{Experimental methodology}
\label{sec:expmethodology}

\subsection{Dividing the data}
\label{section:dividing}
Having a database that gathers various eye pathologies does not necessarily mean that each eye is affected by one illness only. In most cases there are two or even more conditions present in one eye. Some of them do not affect the iris at all, some impact pupillary regions, other target the cornea or the iris tissue itself. This abundance of various and often unrelated medical conditions makes the analysis difficult. Therefore, we come up with a solution of subdividing the dataset into several groups, connected not by illnesses present in the eye themselves, but by the type of impact they have on iris or other structures of the eyeball, regardless of their medical origin.

\begin{figure*}[!ht]
\centering
\includegraphics[width=0.66\textwidth]{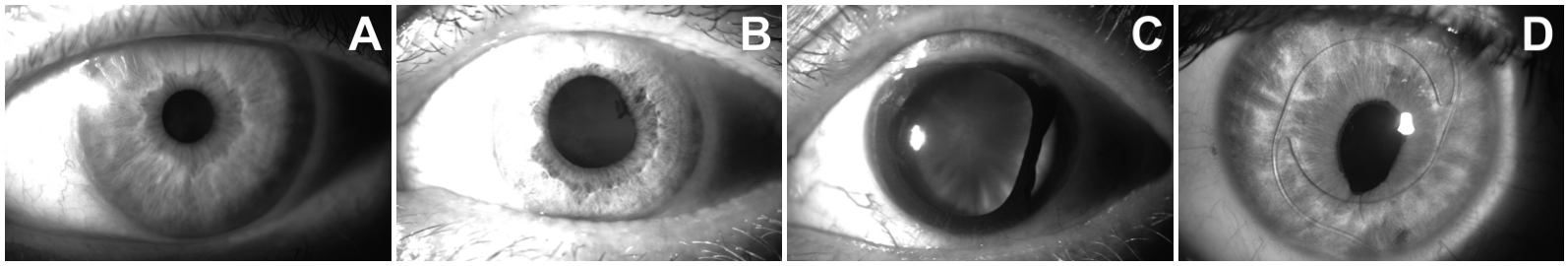}
\caption{Samples representing example ocular conditions that determine inclusion of that particular eye into one of the subsets, namely \emph{Clear} (A), \emph{Geometry} (B), \emph{Tissue} (C) or \emph{Obstructions} (D).}
\label{fig:dividing}
\end{figure*}

Five major subsets can be distinguished on that basis. Healthy eyes are referred to as the \emph{Healthy} subset and serve as a control group. Remaining, disease-affected eyes (identified by an opthalmologist in visible light) are divided into the following groups: no visible changes (\emph{Clear}), eyes with changes in pupillary regions such as deviation from pupil's circularity (\emph{Geometry}), eyes with visible alterations to the iris tissue (\emph{Tissue}) and eyes with the iris covered by obstructions located in front of it (\emph{Obstructions}). See Fig. \ref{fig:dividing} for samples illustrating each subset. Table \ref{database_summary_divided} provides numbers of eyes and samples acquired in NIR illumination in each group. Note that these groups are not disjoint, therefore the total number of eyes and images is larger than the corresponding values from Table \ref{database_summary}.

\begin{table}[!htb]
\renewcommand{\arraystretch}{1.1}

\caption{Database summary in respect to illness impact on certain eye structures (NIR-illuminated samples).}
\label{database_summary_divided}
\centering
\begin{tabular}{|c|c|c|c|}
\hline
\textbf{Influence type}  & \textbf{Subset name} & \textbf{No. of eyes} & \textbf{No. of NIR samples}\\
\hline
\hline
Healthy eyes & \emph{Healthy} & 12  &  93\\
\hline
No visible changes & \emph{Clear} & 46  & 291 \\
\hline
Distorted pupil geometry & \emph{Geometry} & 18 & 196\\
\hline
Iris tissue alterations & \emph{Tissue} & 11 & 87\\
\hline
Obstructed iris &  \emph{Obstructions} & 10 & 70\\
\hline
\hline
\textbf{Total} & -- & \textbf{97} & \textbf{737}\\
\hline
\end{tabular}
\end{table}

\subsection{Similarity scores generation}
For each of the five groups of images discussed in Section \ref{section:dividing} a distribution of similarity scores has been calculated using all three iris recognition methods (Sec. \ref{section:methods}). All possible pairs of genuine (intra-class) comparisons as well as impostor (inter-class) comparisons have been taken into account in this research to estimate how eye illness influence the authentication accuracy. It should be noted that each image pair was used only once, \emph{i.e.}, if sample A was compared to sample B, then the comparison was not repeated in reverse order (sample B against sample A).

We perform statistical analysis on whether the average similarity scores obtained from the genuine and impostor comparisons between samples in subsets \emph{Clear}, \emph{Geometry}, \emph{Tissue} and \emph{Obstructions} are different from those calculated using the control subset \emph{Healthy}. This is to come up with insight if certain types of impact on eye structures cause more erroneous performance of selected recognition methodologies that the other. To answer the question whether a particular disease impacts the iris recognition, for each type of comparisons (genuine and impostor) a one-tailed t-test is carried out with a significance level $\alpha = 0.05$. Cumulative distribution functions are also gathered together and compared with each other to illustrate the scale and direction of possible shifts among these distributions, especially in reference to distributions obtained using \emph{Healthy} eyes subset.

\section{Results}
\label{sec:Results}

\begin{figure}[!htb]
\centering
\begin{subfigure}{0.25\textwidth}
  \centering
  \includegraphics[width=0.99\linewidth]{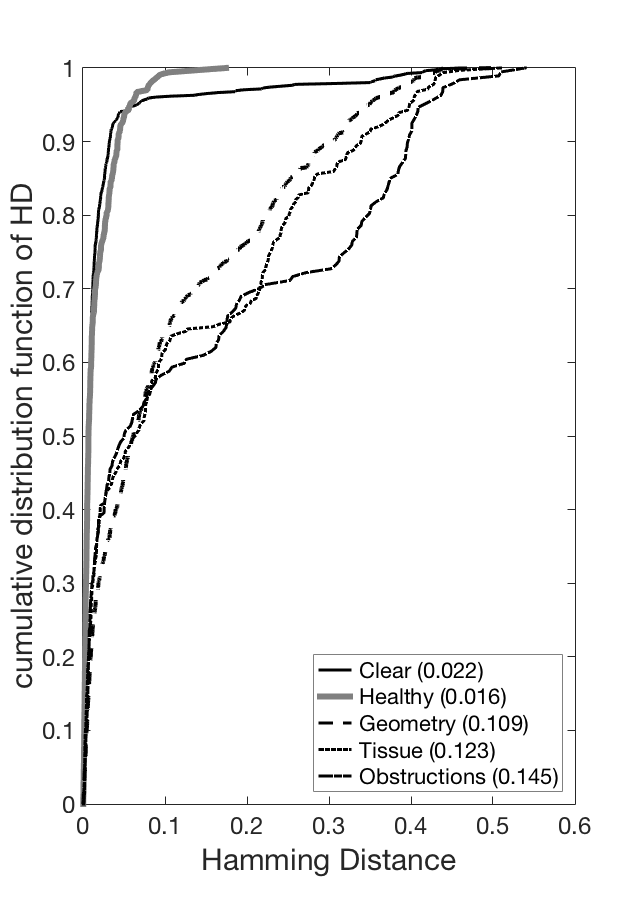}
\end{subfigure}%
\begin{subfigure}{0.25\textwidth}
  \centering
  \includegraphics[width=0.99\linewidth]{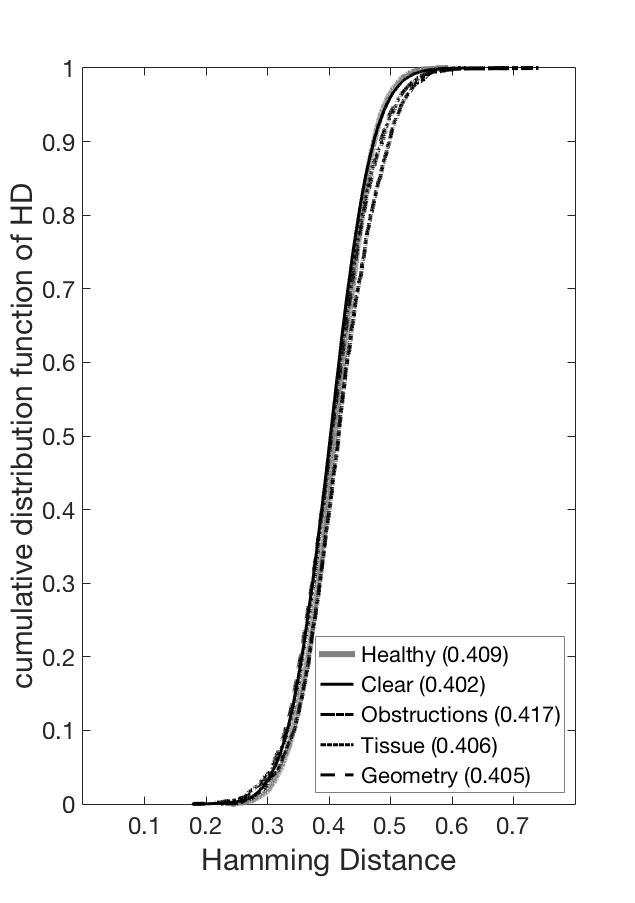}
\end{subfigure}
\caption{Cumulative distributions of genuine (left) and impostor (right) scores shown independently for data gathered in five subsets of eyes: \emph{Healthy, Clear, Geometry, Tissue} and \emph{Obstructions}. Comparison scores are calculated by MIRLIN matcher, \emph{i.e.}, lower score denotes a better match. Average similarity scores for each distribution are also shown in brackets.}
\label{fig:results_mirlin}
\end{figure}

An analysis of differences in average {\bf genuine scores for MIRLIN} method of \emph{Clear, Geometry, Tissue} and \emph{Obstructions} eyes against the \emph{Healthy} subset revealed an increase of the score (\emph{i.e.}, dissimilarity of samples) in all cases by as high as 37\% for \emph{Clear}, 570.6\% for \emph{Geometry}, 659\% for \emph{Tissue} and 792.7\% for \emph{Obstructions}. Fig. \ref{fig:results_mirlin} (left) compares cumulative distributions of genuine comparison scores in all groups, especially illustrates significant deterioration in genuine scores for \emph{Geometry}, \emph{Tissue} and \emph{Obstructions} groups. When looking at changes in average {\bf impostor scores for MIRLIN} method (Fig. \ref{fig:results_mirlin}, right), a decrease in average Hamming distances (\emph{i.e.}, increase in similarity between different irises) is smaller than for genuine ones, namely 1.6\% for \emph{Clear}, 0.9\% for \emph{Geometry} and 0.8\% for \emph{Tissue} groups. When analyzing the \emph{Obstructions} subset, we see an opposite trend, that is an increase in average comparison score by 2.1\% when compared to \emph{Healthy} eyes. All identified differences (for both the genuine and impostor scores) are statistically significant (Table \ref{table:statTests}, rows marked as `MIRLIN'). It leads to a conclusion that even \emph{Clear} eyes (revealing no changes under visible inspection) give worse comparison scores when compared to \emph{Healthy} samples.

\begin{figure}[!htb]
\centering
\begin{subfigure}{0.25\textwidth}
  \centering
  \includegraphics[width=0.99\linewidth]{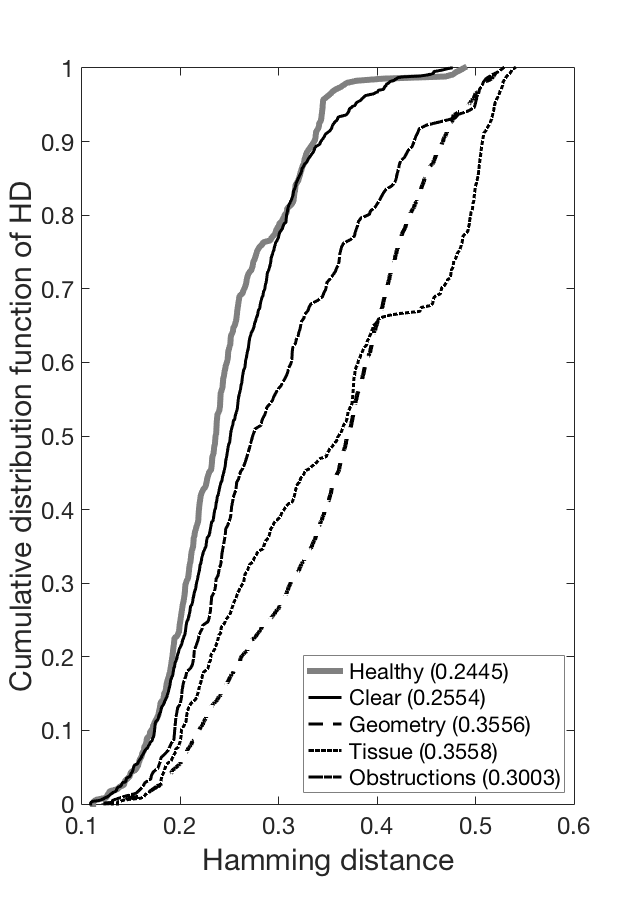}
\end{subfigure}%
\begin{subfigure}{0.25\textwidth}
  \centering
  \includegraphics[width=0.99\linewidth]{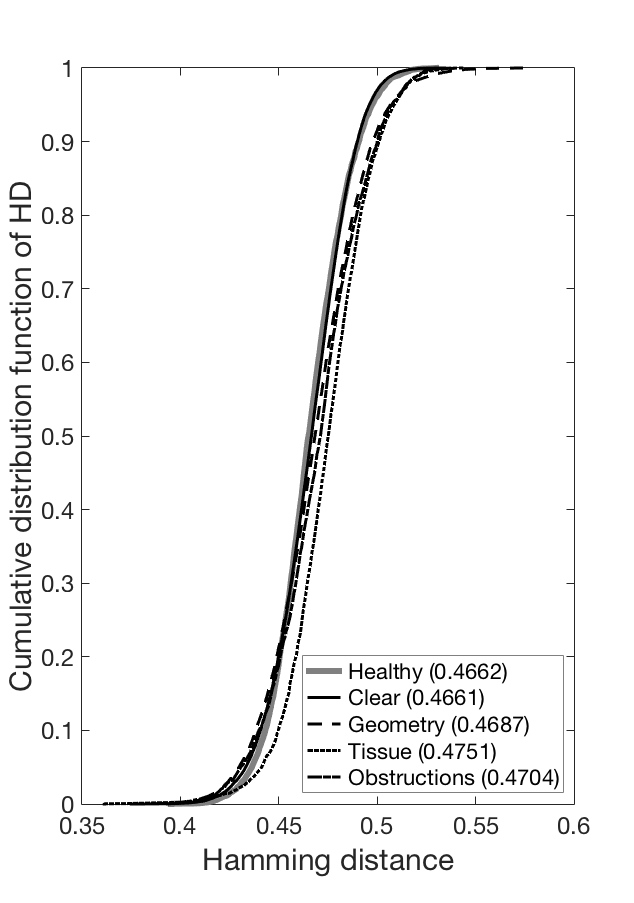}
\end{subfigure}
\caption{Same as in Fig. \ref{fig:results_mirlin}, except OSIRIS method was used to generate comparison scores.}
\label{fig:results_osiris}
\end{figure}

Similarly to the MIRLIN matcher, for {\bf OSIRIS} matcher we observe increase in the average {\bf genuine comparison score} when comparing all four non-healthy eyes against the reference \emph{Healthy} subset: 4.5\% for \emph{Clear} eyes, 45.4\% for the \emph{Geometry} subset, 45.5\% for the \emph{Tissue} subset and 22.9\% for the \emph{Obstructions} subset. Fig. \ref{fig:results_osiris} (left) illustrates cumulative distributions of genuine scores for all five groups. Again, as for the MIRLIN method, differences between \emph{Healthy} and \emph{Clear} eyes are lower, but significant for the remaining, disease-affected irides. All differences are statistically significant (cf. Table \ref{table:statTests}, rows marked as 'OSIRIS', columns marked as 'Genuine comparison scores'). When {\bf OSIRIS impostor score} distributions are involved, we found no statistically significant differences between scores obtained for \emph{Clear} and \emph{Healthy} eyes. However, for subsets \emph{Geometry}, \emph{Tissue} and \emph{Obstructions} the average impostor scores were larger by 0.5\%, 1.9\% and 0.9\%, respectively, when compared with the \emph{Healthy} subset, and these increases are statistically significant (cf. Table \ref{table:statTests}, rows marked as `OSIRIS', columns marked as `Impostor comparison scores'). Fig. \ref{fig:results_osiris} (right) illustrates the cumulative distributions of OSIRIS impostor scores in all five groups.

\begin{figure}[!htb]
\centering
\begin{subfigure}{0.25\textwidth}
  \centering
  \includegraphics[width=0.99\linewidth]{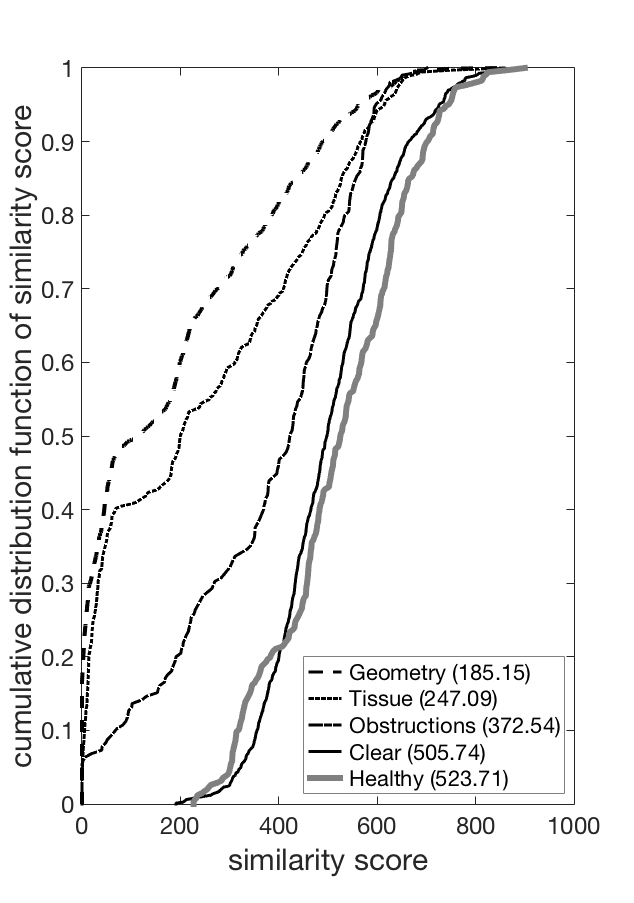}
\end{subfigure}%
\begin{subfigure}{0.25\textwidth}
  \centering
  \includegraphics[width=0.99\linewidth]{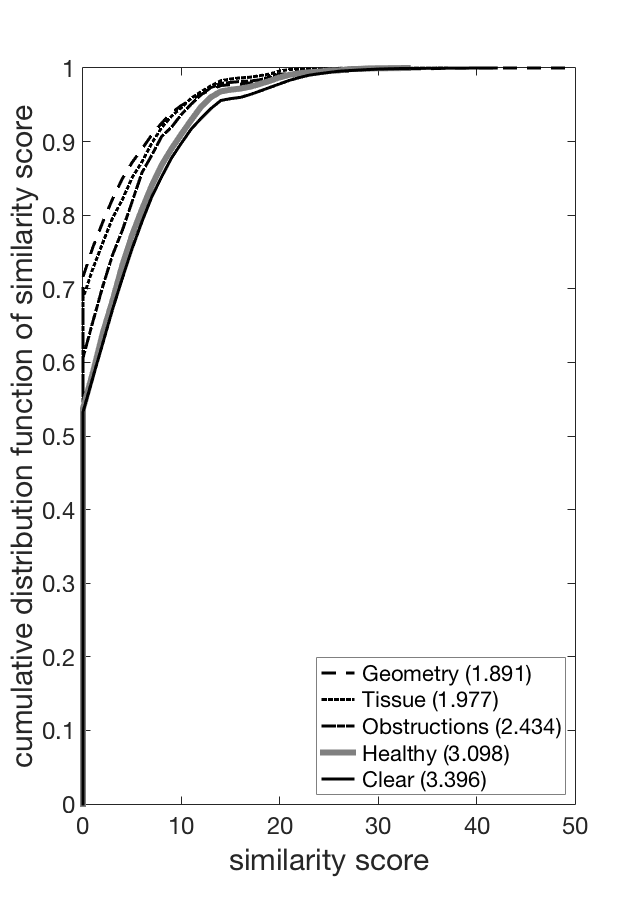}
\end{subfigure}
\caption{Same as in Fig. \ref{fig:results_mirlin}, except VeriEye method was used to generate comparison scores. Note the opposite polarity of comparison scores when compared to MIRLIN and OSIRIS methods: the higher the score, the better the match.}
\label{fig:results_neuro}
\end{figure}

{\bf VeriEye} comparison scores have opposite polarity when compared to the first two methods, namely the higher the score the better the match. It means that mean values of {\bf genuine comparison scores} in four groups of non-healthy eyes are worse (\emph{i.e.}, lower) when compared to \emph{Healthy} subsets. We observe a decrease of average genuine scores by 3.4\% for \emph{Clear} eyes, 64.7\% for \emph{Geometry} subset, 52.8\% for \emph{Tissue} subset and 28.9\% for \emph{Obstructions} group. As for MIRLIN and OSIRIS methods, all differences are statistically significant (cf. Table \ref{table:statTests}, rows marked as `VeriEye', columns marked as `Genuine comparison scores'). Fig. \ref{fig:results_neuro} (left) shows cumulative distributions comparing the scores in all five subsets. What should be noted is an extremely poor performance in the \emph{Geometry} and \emph{Tissue} subsets, and bad performance in the \emph{Obstructions} subset. In particular, in all three subsets we get comparison scores as low as 0 (values below 30 are classified as different-eye scores according to the default VeriEye matching rules) and therefore would generate false non-matches. Comparison of the {\bf VeriEye impostor average scores} against the \emph{Healthy} subset average scores revealed an increase by 9.6\% for \emph{Clear} subset, and a decrease for the remaining three subsets: by 189.1\% for \emph{Geometry}, 36.2\% for \emph{Tissue} and 21.4\% for \emph{Obstructions} subsets. There is little difference between all five subsets (cf. Fig. \ref{fig:results_neuro}, right), however, for subsets \emph{Clear}, \emph{Geometry}, \emph{Tissue} and \emph{Healthy} there are values reaching beyond 30, and thus those pairs of samples would be classified as same-eye images (\emph{i.e.}, we have a risk of false matches). All differences are statistically significant (see Table \ref{table:statTests}, rows marked as `VeriEye', columns marked as `Impostor comparison scores').

\newlength{\colw}
\setlength{\colw}{1.1cm}

\begin{table*}[!t]
\renewcommand{\arraystretch}{1.1}

\caption{Summary of statistical testing for three different iris recognition methods related to differences in average comparison scores obtained in different groups of eye diseases. All tests use one-tailed t-test at significance level $\alpha = 0.05$. In all tests the null hypothesis H0 states that the scores from two subsets being compared (for instance, \emph{Healthy} and \emph{Clear}) come from independent random samples with equal means and equal, but unknown variances. Alternative hypotheses H1 are defined as in the rows labeled `H1'. Corresponding $p$-values that are shown for each test suggest that in all cases but one (OSIRIS impostor scores for \emph{Healthy} and \emph{Clear} eyes) the null hypothesis should be rejected. It means that differences in mean comparison scores among these groups are statistically significant.}
\label{table:statTests}
\centering
\begin{tabular}{cc|>{\centering}m{\colw}|>{\centering}m{\colw}|>{\centering}m{\colw}|>{\centering}m{\colw}|c|%
|>{\centering}m{\colw}|>{\centering}m{\colw}|>{\centering}m{\colw}|>{\centering}m{\colw}|c|}
\cline{3-12}
& & \multicolumn{5}{c||}{\bf Genuine comparison scores} & \multicolumn{5}{c|}{\bf Impostor comparison scores} \\
\cline{3-12}
& & \emph{Healthy} & \emph{Clear}  & \emph{Geometry} & \emph{Tissue} & \emph{Obstructions}  & \emph{Healthy} & \emph{Clear}  & \emph{Geometry} & \emph{Tissue} & \emph{Obstructions} \\
& & ($g_h$) & ($g_c$)  & ($g_g$) & ($g_t$) & ($g_o$) & ($i_h$) & ($i_c$)  & ($i_g$) & ($i_t$) & ($i_o$) \\\hline\hline
\multicolumn{1}{|c}{\multirow{3}{*}{MIRLIN}} & \multicolumn{1}{|c|}{\bf mean} & 0.0162 & 0.0222 & 0.1086 & 0.1229 & 0.1446 & 0.4089 & 0.4023& 0.4052 & 0.4058 & 0.4173 \\\cline{2-12}
\multicolumn{1}{|c}{} & \multicolumn{1}{|c|}{\bf H1} & & $\bar{g}_c > \bar{g}_h$ & $\bar{g}_g > \bar{g}_h$ & $\bar{g}_t > \bar{g}_h$ & $\bar{g}_o > \bar{g}_h$ & & $\bar{i}_c < \bar{i}_h$ & $\bar{i}_g < \bar{i}_h$ & $\bar{i}_t < \bar{i}_h$ & $\bar{i}_o > \bar{i}_h$ \\\cline{2-12}
\multicolumn{1}{|c}{} & \multicolumn{1}{|c|}{\bf $p$-value} & & 0.0416 & \texttildelow 0 &  \texttildelow 0 &  \texttildelow 0 & & \texttildelow 0 & 0.0004 & 0.0144 & \texttildelow 0 \\\hline\hline
\multicolumn{1}{|c}{\multirow{3}{*}{OSIRIS}} & \multicolumn{1}{|c|}{\bf mean} & 0.2445  &  0.2554  &  0.3556  &  0.3558  &  0.3003 & 0.4662 &    0.4661  &   0.4687  &   0.4751  &   0.4704 \\\cline{2-12}
\multicolumn{1}{|c}{} & \multicolumn{1}{|c|}{\bf H1} & & $\bar{g}_c > \bar{g}_h$ & $\bar{g}_g > \bar{g}_h$ & $\bar{g}_t > \bar{g}_h$ & $\bar{g}_o > \bar{g}_h$ & & $\bar{i}_c < \bar{i}_h$ & $\bar{i}_g > \bar{i}_h$ & $\bar{i}_t > \bar{i}_h$ & $\bar{i}_o > \bar{i}_h$ \\\cline{2-12}
\multicolumn{1}{|c}{} & \multicolumn{1}{|c|}{\bf $p$-value} & & 0.005 & \texttildelow 0  & \texttildelow 0  & \texttildelow 0  & & 0.3756 & \texttildelow 0 & \texttildelow 0 & \texttildelow 0 \\\hline\hline
\multicolumn{1}{|c}{\multirow{3}{*}{VeriEye}} & \multicolumn{1}{|c|}{\bf mean} & 523.71 & 505.74  &185.15 & 247.09 & 372.54 & 3.098  &  3.396 & 1.891  & 1.977 & 2.434 \\\cline{2-12}
\multicolumn{1}{|c}{} & \multicolumn{1}{|c|}{\bf H1} & & $\bar{g}_c < \bar{g}_h$ & $\bar{g}_g < \bar{g}_h$ & $\bar{g}_t < \bar{g}_h$ & $\bar{g}_o < \bar{g}_h$ & & $\bar{i}_c > \bar{i}_h$ & $\bar{i}_g < \bar{i}_h$ & $\bar{i}_t < \bar{i}_h$ & $\bar{i}_o < \bar{i}_h$ \\\cline{2-12}
\multicolumn{1}{|c}{} & \multicolumn{1}{|c|}{\bf $p$-value} & & 0.0137 & \texttildelow 0 & \texttildelow 0 & \texttildelow 0 & & 0.0004 & \texttildelow 0 & \texttildelow 0 & \texttildelow 0 \\\hline
\end{tabular}
\end{table*}

\section{Conclusions}

This paper delivers a few important observations that are interesting in various ways. First, it is rather uncommon in operational practice to see non-healthy eyes that are affected by a single disease only. It means that the analysis of how the performance of iris recognition methods is affected by eye conditions should not include only single pathologies in the employed dataset. This is why our database was divided into five groups relating to the expected impact on the iris recognition accuracy. Second, the statistically significant differences in comparison scores for healthy eyes and non-healthy, yet invoking no suspicions when near-infrared samples are visually inspected, suggest that changes in iris texture and/or geometry invisible to the biometric expert can still deteriorate the iris recognition reliability. Third, the deterioration in genuine scores is much worse when healthy eyes are compared with those affected by diseases causing visible pathologies in the iris appearance. Fourth, although the differences in average impostor comparison scores are statistically significant (as for genuine scores) they do not significantly increase the risk of false matches in all iris recognition methods applied in this research. The database is publicly available to the  biometric research community for non-commercial purposes.

\bibliographystyle{IEEEtran}
\bibliography{refs}

\end{document}